\documentclass[sigconf]{acmart}

\usepackage{amsmath}
\usepackage{mathtools}
\usepackage{amsthm}
\usepackage{caption}
\usepackage{microtype}
\usepackage{graphicx}
\usepackage{booktabs}
\usepackage{hyperref}
\usepackage{enumitem}
\usepackage{multirow}
\usepackage{multicol}
\usepackage{ulem}
\usepackage{algorithm}
\usepackage{algorithmic}
\usepackage{tabularx}
\usepackage{subcaption}

\usepackage{placeins}
\usepackage[table]{xcolor}
\usepackage{amsmath} 
\usepackage{bm}     

\AtBeginDocument{%
  }

\setcopyright{acmlicensed}
\copyrightyear{2018}
\acmYear{2018}
\acmDOI{XXXXXXX.XXXXXXX}

\acmConference[Conference acronym 'XX]{Make sure to enter the correct
  conference title from your rights confirmation emai}{June 03--05,
  2018}{Woodstock, NY}
\acmISBN{978-1-4503-XXXX-X/18/06}

\begin{document}

\title{DSO: Dual-Scale Neural Operators for Long-term Fluid Dynamics Forecasting}
\def\method{method}

\author{Huanshuo Dong}
\authornote{Equal contribution.}
\email{bingo000@mail.ustc.edu.cn}
\affiliation{%
  \institution{University of Science and Technology of China}
  \city{Hefei}
  \country{China}
}

\author{Hao Wu}
\authornotemark[1]
\affiliation{%
  \institution{Tsinghua University}
  \city{Beijing}
  \country{China}
}

\author{Hong Wang}
\authornote{Corresponding author.}
\email{wanghong1700@mail.ustc.edu.cn}
\affiliation{%
  \institution{University of Science and Technology of China}
  \city{Hefei}
  \country{China}
}

\author{Qin-Yi Zhang}
\affiliation{%
  \institution{Institute of Automation, Chinese Academy of Sciences}
  \city{Beijing}
  \country{China}
}

\author{Zhezheng Hao}
\affiliation{%
  \institution{Zhejiang University}
  \city{Hangzhou}
  \country{China}
}

\renewcommand{\shortauthors}{Dong and Wu, et al.}

\begin{abstract}

Long-term fluid dynamics forecasting is a critically important problem in science and engineering. While neural operators have emerged as a promising paradigm for modeling systems governed by partial differential equations (PDEs), they often struggle with long-term stability and precision. We identify two fundamental failure modes in existing architectures: (1) local detail blurring, where fine-scale structures such as vortex cores and sharp gradients are progressively smoothed, and (2) global trend deviation, where the overall motion trajectory drifts from the ground truth during extended rollouts.
We argue that these failures arise because existing neural operators treat local and global information processing uniformly, despite their inherently different evolution characteristics in physical systems. To bridge this gap, we propose the Dual-Scale Neural Operator (DSO), which explicitly decouples information processing into two complementary modules: depthwise separable convolutions for fine-grained local feature extraction and an MLP-Mixer for long-range global aggregation.
Through numerical experiments on vortex dynamics, we demonstrate that nearby perturbations primarily affect local vortex structure while distant perturbations influence global motion trends—providing empirical validation for our design choice. Extensive experiments on turbulent flow benchmarks show that DSO achieves state-of-the-art accuracy while maintaining robust long-term stability, reducing prediction error by over 88\% compared to existing neural operators.
\end{abstract}

\vspace{-10pt}


\maketitle

\section{Introduction}
\label{sec:intro}

\begin{figure}[t]
\centering
\begin{subfigure}{0.5\textwidth}
    \centering
    \includegraphics[width=\linewidth]{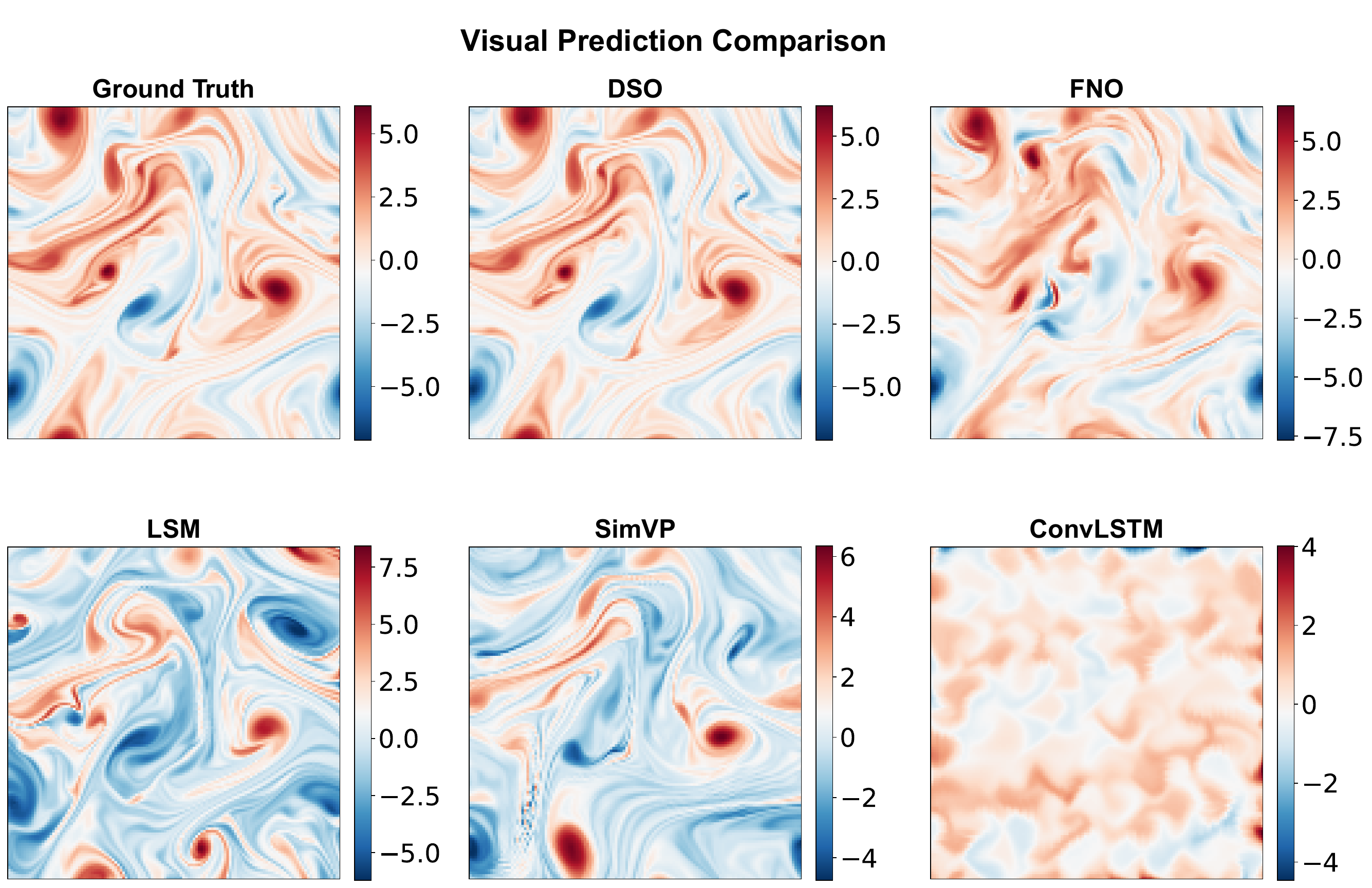}
    \caption{Visualization of long-term predictions for turbulence datasets from different models, where Ground Truth is the actual value and DSO is our method.}
    \label{fig:intro1}
\end{subfigure}
\hfill
\begin{subfigure}{0.48\textwidth}
    \centering
    \includegraphics[width=\linewidth]{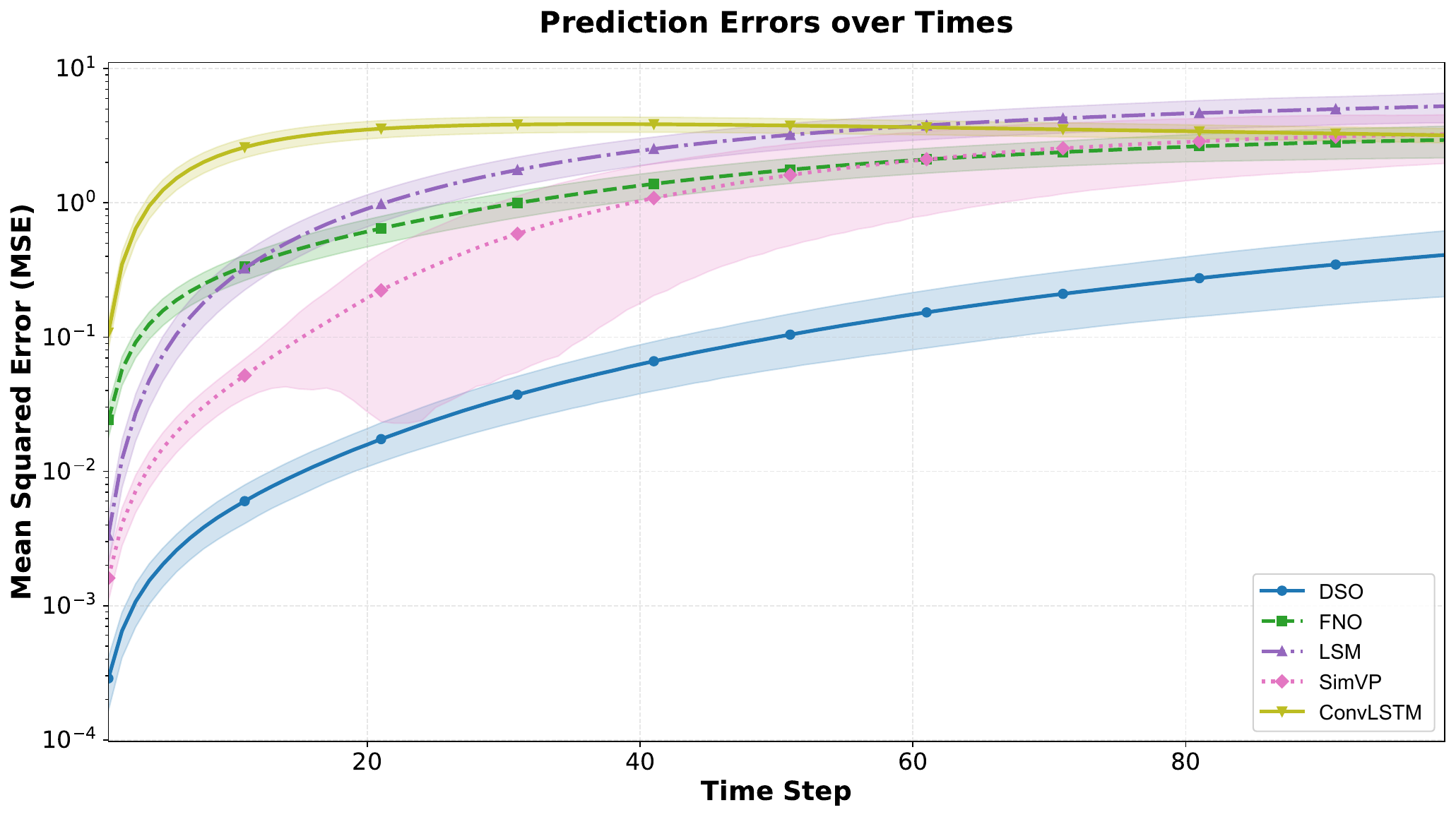}
    \caption{Prediction errors of different models over time through autoregressive rollout. The shaded areas represent the variance across different samples.}
    \label{fig:intro2}
\end{subfigure}
\vspace{-0.5cm}
\label{fig:intro}
\end{figure}

Governed by partial differential equations (PDEs), spatiotemporal dynamics prediction is a fundamental and critically important area spanning physics, mathematics, and engineering disciplines \cite{kovachki2024operator}. Accurate long-term prediction of such dynamics is essential for scientific discovery and technological advancement, underpinning applications that range from weather forecasting and climate modeling to computational fluid dynamics \cite{palmer2019stochastic, duraisamy2019turbulence, norton2006computational}. In recent years, this field has experienced a paradigm shift with the emergence of neural operators. These advanced deep learning architectures are explicitly designed to learn the latent solution operator—the mapping between infinite-dimensional function spaces for Partial Differential Equations (PDEs). Pioneering works, such as the Fourier Neural Operator (FNO) \cite{li2021fourier}, the Convolutional Neural Operator (CNO) \cite{raonic2023convolutional}, and Latent Spectral Models (LSM) \cite{pmlr-v202-wu23f}, have demonstrated remarkable success in solving various PDEs, offering significant computational speedups over traditional numerical solvers while maintaining reasonable accuracy for short-term predictions.

Although neural operators have achieved initial success in short-term prediction, applying them to long-term autoregressive forecasting—where the model's output at one time step becomes the input for the next—still faces significant challenges. As shown in Figure~\ref{fig:intro2}, existing methods exhibit a sharp increase in prediction error over extended time horizons. Figure~\ref{fig:intro1} reveals two characteristic failure modes that compound over time. \textbf{(1) Local Detail Blurring}: Fine-scale structures such as vortex cores, sharp gradients, and turbulent eddies are progressively smoothed out. This phenomenon, analogous to numerical diffusion in classical solvers, results in loss of physically important small-scale features. Methods like FNO \cite{li2021fourier} produce increasingly blurred predictions, failing to preserve the sharp boundaries of coherent structures. \textbf{(2) Global Trend Deviation}: The overall motion trajectory—including vortex translation, rotation, and large-scale flow patterns—gradually drifts from the ground truth. This manifests as phase errors where predicted structures appear at wrong spatial locations, even when local features are reasonably preserved. For instance, SimVP \cite{gao2022simvp} maintains local sharpness but exhibits significant positional drift over extended rollouts.

We attribute these failures to a fundamental limitation: \textit{existing neural operators treat local and global information processing uniformly}, despite their inherently different characteristics in physical systems. In fluid dynamics, nearby interactions directly affect local vortex structure through mechanisms like vortex merging and strain-induced deformation, while distant interactions influence global motion trends through pressure-mediated coupling that acts across the entire domain. A single computational mechanism cannot optimally capture both effects. To validate this hypothesis, we conduct numerical experiments with vortex pairs at varying separations (Section~\ref{sec:motivation}). Our results demonstrate a clear dichotomy: closely-spaced vortices exhibit strong local deformation with minimal center displacement, while widely-separated vortices show little local interaction but significant trajectory deflection. This empirical evidence motivates explicit separation of local and global processing pathways.

To overcome this critical limitation, we propose the \textbf{Dual-Scale Neural Operator (DSO)}, which introduces a physics-motivated dual-pathway mechanism for spatiotemporal data processing. Our architecture integrates two synergistic components: (1) a \textbf{Local Processing Module via Depthwise Separable Convolution} that captures fine-scale features, gradient structures, and local interactions within a bounded receptive field, responsible for preserving local details and preventing numerical diffusion; and (2) a \textbf{Global Processing Module via MLP-Mixer} that aggregates information across the entire spatial domain through channel-wise and spatial mixing operations, capturing long-range dependencies and maintaining correct global motion trends. By explicitly decoupling these two processing pathways, DSO simultaneously prevents local detail blurring and global trend deviation—addressing both failure modes of existing approaches.

Extensive validation on canonical turbulent flow benchmarks unequivocally demonstrates the superior performance and robust stability of DSO in long-term spatiotemporal forecasting. Our method achieves substantial improvement in prediction accuracy, reducing prediction error by over 88\% compared to existing state-of-the-art methods. Crucially, DSO maintains robust stability over extended prediction horizons where conventional methods quickly fail. This stability stems from the dual-pathway mechanism's ability to effectively address both local diffusion and global drift simultaneously. Our main contributions are: 
\begin{itemize}[leftmargin=*,nosep]
    \item We identify and analyze two fundamental failure modes in long-term neural operator predictions: local detail blurring and global trend deviation.
    
    \item Through numerical experiments on vortex dynamics, we demonstrate that local and global perturbations have qualitatively different effects, providing physical motivation for separate processing mechanisms.
    
    \item We propose DSO, a dual-scale neural operator that explicitly decouples local and global information processing through convolution and MLP-Mixer modules.
    
    \item Extensive experiments demonstrate that DSO achieves state-of-the-art accuracy while maintaining long-term stability across multiple challenging benchmarks.
\end{itemize}

\section{Related Work}
\label{sec:Related Work}
\subsection{Turbulence prediction}
Turbulence prediction remains a challenging problem due to the complex, chaotic nature of fluid flows governed by the Navier-Stokes (NS) equations \cite{landau1987fluid}. Traditional numerical methods, such as direct numerical simulation and large eddy simulation, provide accurate solutions but suffer from high computational costs, particularly for long-term predictions \cite{pope2001turbulent}. Data-driven approaches, such as deep learning-based models, have emerged as promising alternatives, offering potential for real-time forecasting with reduced computational overhead. Early works like \cite{srinivasan2019predictions} applied neural networks to predict turbulent dynamics, but these models struggled with generalizing to unseen flow regimes. Recent advancements have incorporated physical constraints, like the NS equations, into neural network training, ensuring better adherence to the governing laws of fluid dynamics \cite{wang2020towards, kag2022physics}. Despite these improvements, long-term prediction of turbulence still faces issues like loss of fine-scale features and failure to preserve physical consistency in extended forecasts \cite{cai2024temporally}. These challenges highlight the need for more sophisticated models that can integrate spatial and temporal adaptivity, as well as global physical consistency.

\subsection{Neural operator and video prediction model}
Neural operators have become central to solving complex PDEs by directly learning mappings between input and output data. Fourier Neural Operator (FNO) \cite{li2021fourier} leverages the Fast Fourier Transform to efficiently approximate solutions in the spectral domain, balancing accuracy with computational efficiency. Further extensions, such as U-NO \cite{rahman2022u} and CNO \cite{raonic2023convolutional}, enhance the model’s ability to handle multi-scale dynamics. LSM \cite{pmlr-v202-wu23f} addresses high-dimensional PDEs by incorporating spectral methods in learned latent spaces. Models such as ConvLSTM \cite{lin2020self}, ResNet \cite{he2016deep}, and PastNet \cite{wu2024pastnet} excel in capturing temporal dependencies, making them well-suited for dynamic forecasting tasks like turbulence prediction. The integration of attention mechanisms in models like Swin Transformer \cite{liu2021swin} and SimVP \cite{gao2022simvp} further enhances their ability to model long-range temporal dynamics. These models offer promising avenues for advancing turbulence prediction, where both spatial and temporal complexities must be captured simultaneously.

\section{Preliminary}
\label{sec:Preliminary}
\subsection{Problem Definition}

We consider the two-dimensional (2D) Navier-Stokes equations for a viscous, incompressible fluid \cite{li2021fourier, takamoto2022pdebench, evans2022partial}. The dynamics are expressed in vorticity form on the unit torus $\mathbb{T}^2 = [0, 2\pi]^2$ with periodic boundary conditions. The evolution of the scalar vorticity field, $\boldsymbol{\omega}(x,t)$, is governed by:
\begin{align}
    \partial_t \boldsymbol{\omega} + (\mathbf{u} \cdot \nabla) \boldsymbol{\omega} &= \nu \Delta \boldsymbol{\omega} + f, &&(x,t) \in \mathbb{T}^2 \times (0,T], \label{eq:vorticity} \\
    \boldsymbol{\omega}(x,0) &= \boldsymbol{\omega}_0(x), && x \in \mathbb{T}^2.
\end{align}
Here, $\boldsymbol{\omega}$ is the vorticity, $\mathbf{u}$ is the divergence-free velocity field, $\nu > 0$ is the kinematic viscosity, and $f$ is a time-independent external forcing. The velocity field $\mathbf{u}$ is recovered from the vorticity $\boldsymbol{\omega}$ via the Biot-Savart law, which is defined through the stream-function $\psi$:
\begin{equation}
\mathbf{u} = \nabla^\perp \psi = (\partial_{x_2} \psi, -\partial_{x_1} \psi), \quad  -\Delta \psi = \boldsymbol{\omega}.
\end{equation}
This relationship closes the system, rendering the advection term $(\mathbf{u} \cdot \nabla) \boldsymbol{\omega}$ a non-local quadratic nonlinearity in $\boldsymbol{\omega}$. For suitable initial conditions $\boldsymbol{\omega}_0$ and forcing $f$ (e.g., in $L^2(\mathbb{T}^2)$), the system is globally well-posed, guaranteeing a unique solution trajectory whose regularity depends on the smoothness of the initial data \cite{evans2022partial}.

Our objective is to learn a parameterized operator, $\mathcal{G}_\theta$, that approximates the true solution operator, $\mathcal{G}$ \cite{li2021fourier}. This operator maps a history of the vorticity field over an input interval $[0, T_{in}]$ to its future evolution over a subsequent interval $(T_{in}, T]$. Formally, the mapping is:
\begin{equation}
\mathcal{G}: C([0, T_{in}]; H^r(\mathbb{T}^2)) \to C((T_{in}, T]; H^r(\mathbb{T}^2)),
\end{equation}
where $H^r(\mathbb{T}^2)$ is the Sobolev space of order $r$, chosen to reflect the regularity of the data in our experiments. In practice, our model learns a one-step mapping that is applied autoregressively to perform this long-horizon forecasting, effectively propagating the solution in time.

\subsection{Autoregressive Rollout}

The learned operator $\mathcal{G}_\theta$ is trained to predict a single future state from a sequence of past observations. Specifically, given a history of $T_{in}$ consecutive vorticity fields $\{\boldsymbol{\omega}^{(t-T_{in}+1)}, \ldots, \boldsymbol{\omega}^{(t)}\}$, the model predicts the next state:
\begin{equation}
\hat{\boldsymbol{\omega}}^{(t+1)} = \mathcal{G}_\theta(\boldsymbol{\omega}^{(t-T_{in}+1)}, \ldots, \boldsymbol{\omega}^{(t)}).
\end{equation}

For long-term prediction over $T_{pred}$ steps, we apply the model autoregressively. The predicted state $\hat{\boldsymbol{\omega}}^{(t+1)}$ is appended to the input sequence, and the oldest frame is discarded:
\begin{equation}
\hat{\boldsymbol{\omega}}^{(t+k+1)} = \mathcal{G}_\theta(\hat{\boldsymbol{\omega}}^{(t+k-T_{in}+1)}, \ldots, \hat{\boldsymbol{\omega}}^{(t+k)}), \quad k = 1, 2, \ldots, T_{pred}-1.
\end{equation}

Note that after the initial prediction, all subsequent inputs consist entirely of model-generated states, making the model highly susceptible to error accumulation. Any prediction error at step $k$ propagates to all future steps, potentially leading to divergence from the true trajectory. This \textit{error compounding} phenomenon is the central challenge addressed in this work.

\section{Motivation}
\label{sec:motivation}

\begin{figure*}[h]
    \centering
    \includegraphics[width=1\textwidth]{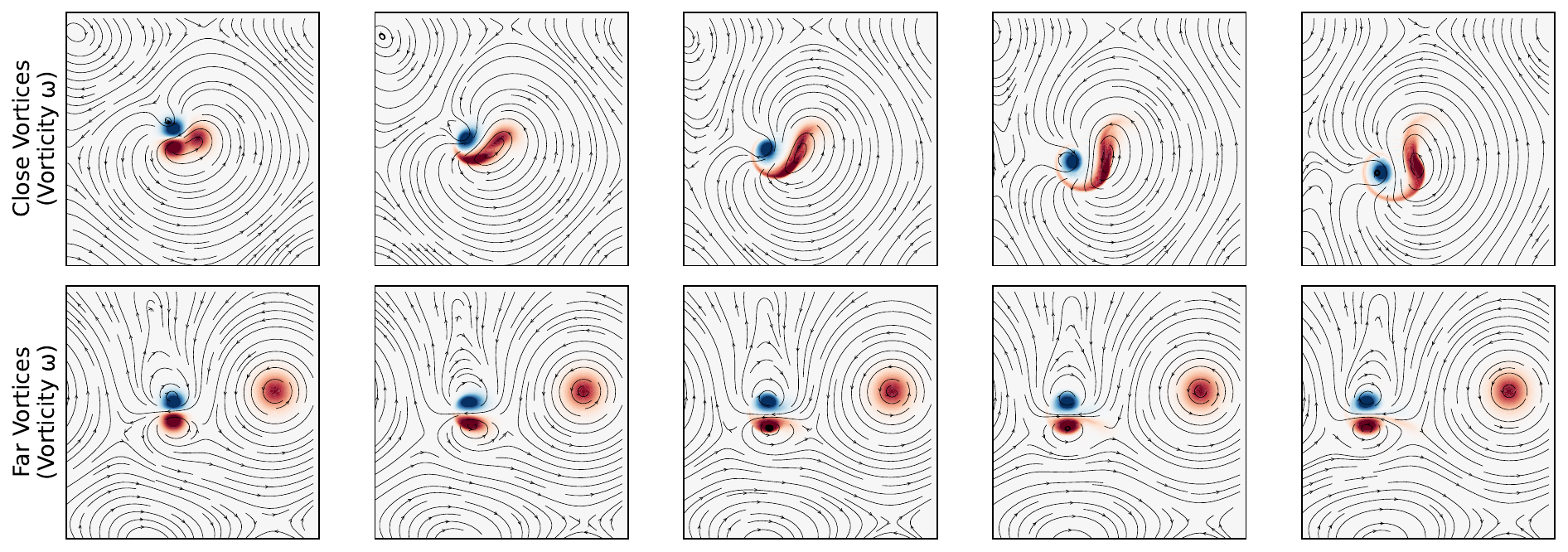}
    \caption{Temporal evolution of vortex dipole under close (top) vs. far (bottom) perturbations. The curve represents a streamline of the flow field. The red-blue vortex pair on the left represents the dipole, while the isolated vortex on the right is the perturbing vortex.}
    \label{fig:motivation}
\end{figure*}

We observe that local and global perturbations have fundamentally different effects on vortex dynamics, indicating that a single-scale mechanism cannot optimally capture both fine-scale structural changes and large-scale trajectory evolution. This observation motivates our dual-scale algorithm design.

\subsection{Experimental Setup}

We simulate 2D incompressible Navier-Stokes flow using a pseudo-spectral method on a $128 \times 128$ periodic domain $[0, 2\pi]^2$ (details in Appendix~\ref{app:motivation_details}). We compare two scenarios: a vortex dipole with a perturbing vortex placed at \textbf{close distance} ($d = 0.6$) vs. \textbf{far distance} ($d = 2.5$), isolating local and global effects respectively.

\subsection{Key Observations}

Figure~\ref{fig:motivation} shows the temporal evolution of both scenarios. We observe strikingly different behaviors:

\textbf{Close Perturbation (Local Effect):} The nearby vortex induces strong \textit{local deformation} of the dipole structure. The vortices stretch, merge partially, and exhibit complex fine-scale interactions. This leads to significant gradient intensification within the local region. Although the system also exhibits global displacement, the dominant effect is the enhancement of fine-scale structure complexity.

\textbf{Far Perturbation (Global Effect):} The distant vortex barely affects the dipole's internal structure. Without nearby disturbance, the local gradients gradually weaken over time. Instead, the far vortex changes the dipole's moving direction through long-range pressure effects, while leaving its internal shape unchanged.

\subsection{Quantitative Analysis}

To quantify these observations, we track two metrics:
\begin{itemize}[leftmargin=*,nosep]
    \item \textbf{Local Deformation}: Maximum vorticity gradient $\max(\|\nabla \omega\|)$. The vorticity gradient measures how rapidly the rotation intensity changes across space. Its increase indicates the formation of fine-scale structures (e.g., sharp vortex edges), while its decrease reflects structural smoothing.
    \item \textbf{Global Displacement}: Center-of-vorticity position 
    $$\mathbf{x}_c = \int \mathbf{x} |\omega| d\mathbf{x} \bigg/ \int |\omega| d\mathbf{x}.$$
\end{itemize}

\begin{small}
\begin{table}[h]
\centering
\caption{Comparison of effects for perturbations.}
\label{tab:motivation}
\begin{tabular}{lcc}
\toprule
Metric & Close ($d=0.6$) & Far ($d=2.5$) \\
\midrule
Local gradient change $\Delta \max(\|\nabla \omega\|)$ & \textbf{+45\%} & $-$29\% \\
Global position shift $\|\Delta \mathbf{x}_c\|$ & 0.8 & 0.6 \\
\bottomrule
\end{tabular}
\end{table}
\end{small}

Table~\ref{tab:motivation} summarizes the results. The key distinction lies in the \textit{nature} of the effect rather than its magnitude. The close perturbation causes a 45\% \textit{increase} in local gradient intensity due to vortex stretching and fine-scale interactions—the nearby vortex actively enhances structural complexity. In contrast, the far perturbation leads to gradient \textit{decay} ($-$29\%) as the dipole evolves undisturbed by local interference; the isolated dipole simply dissipates gradually. While both scenarios show similar global displacement magnitudes (0.8 vs 0.6), the underlying mechanisms are fundamentally different: close perturbations create local structural changes, while far perturbations influence dynamics through non-local pressure coupling without affecting local structure.


This experiment reveals a fundamental dichotomy in fluid dynamics:
\begin{itemize}[leftmargin=*,nosep]
    \item \textbf{Local interactions} (nearby vortices) primarily affect \textit{fine-scale structure} within a bounded spatial region.
    \item \textbf{Global interactions} (distant vortices) primarily affect \textit{motion trajectory} through domain-wide pressure coupling.
\end{itemize}


\section{Method}
\label{sec:method}

Based on our observation that local and global information propagation have fundamentally different effects in fluid dynamics (Section~\ref{sec:motivation}), we propose the \textbf{Dual-Scale Operator (DSO)}. DSO explicitly separates local and global processing through two complementary pathways: convolutions for local feature extraction and MLP-Mixer for global information aggregation.

\begin{figure*}[htbp]
\centering
\includegraphics[width=1\textwidth]{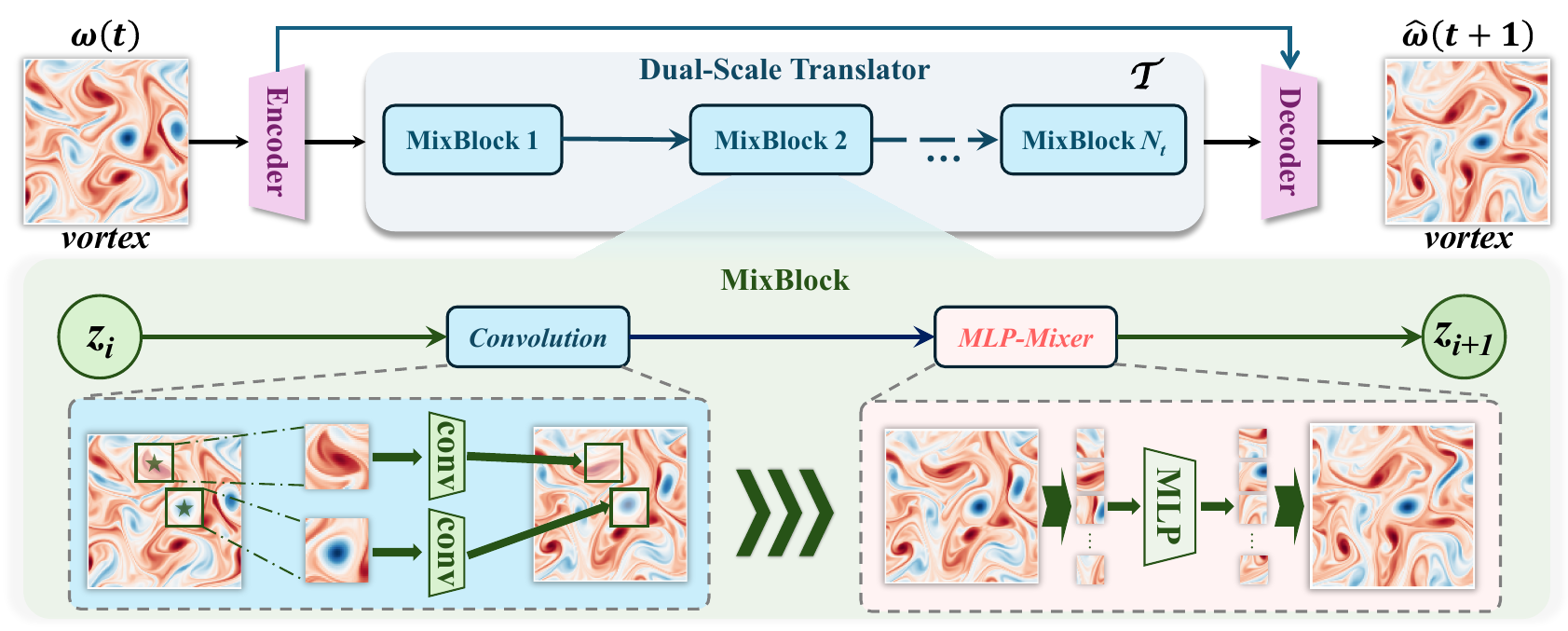}
\caption{Overview of DSO architecture. The model consists of three components: (1) an encoder $\mathcal{E}$ that extracts multi-scale spatial features, (2) a translator $\mathcal{T}$ composed of stacked dual-pathway blocks, and (3) a decoder $\mathcal{D}$ that reconstructs the predicted field. Each block contains a \textbf{Local Pathway} (convolution) for fine-scale features and a \textbf{Global Pathway} (MLP-Mixer) for domain-wide information aggregation. The curved arrows connecting the encoder and decoder represent skip connections.}
\label{fig:method}
\end{figure*}

\subsection{Overall Architecture}

As shown in Figure~\ref{fig:method}, DSO implements the prediction mapping through:
\begin{equation}
\hat{\boldsymbol{\omega}}{(t+1)} = \mathcal{D} \circ \mathcal{T} \circ \mathcal{E}(\boldsymbol{\omega}{(t)}),
\end{equation}
where $\mathcal{E}$, $\mathcal{T}$, and $\mathcal{D}$ represent the encoder, dual-scale translator, and decoder, respectively. The input is a sequence of vorticity fields $\boldsymbol{\omega}_{\text{in}} \in \mathbb{R}^{B \times T \times C \times H \times W}$(batch, time, channel, height, width). The output is the predicted next state $\hat{\boldsymbol{\omega}} \in \mathbb{R}^{B \times T \times C \times H \times W}$.

\subsection{Encoder-Decoder}

\paragraph{Encoder.} The encoder $\mathcal{E}$ progressively reduces spatial resolution while increasing feature channels through $N_s$ convolutional layers:
\begin{equation}
p_i = \sigma(\text{Norm}(\text{Conv}(p_{i-1}))), \quad 1 \leq i \leq N_s,
\end{equation}
where $p_0 = \boldsymbol{\omega}^{(t)}$ is the input vorticity field, $N_s$ is the number of encoder layers, and $\sigma$ is a nonlinear activation. The shallow-layer features $p_1$ are preserved for skip connections.

\paragraph{Decoder.} The decoder $\mathcal{D}$ reconstructs the output through $N_s$ transposed convolutional layers:
\begin{equation}
q_i = \sigma(\text{Norm}(\text{ConvT}(q_{i-1}))), \quad 1 \leq i \leq N_s - 1,
\end{equation}
where $q_0 = \tilde{z}$ is the translator output. The final layer incorporates skip connections from the encoder:
\begin{equation}
\hat{\boldsymbol{\omega}}{(t+1)} = \text{Proj}(\sigma(\text{Norm}(\text{ConvT}([q_{N_s-1}, p_1])))),
\end{equation}
where $[\cdot, \cdot]$ denotes concatenation and $\text{Proj}$ is a $1 \times 1$ convolution that projects to output channels.

\subsection{Dual-Scale Translator}

The translator $\mathcal{T}$ is the core of DSO, consisting of $N_t$ stacked dual-pathway blocks. Each block \textbf{sequentially} applies local and global processing:
\begin{equation}
z' = \mathcal{F}_{\text{local}}(z_k), \quad z_{k+1} = \mathcal{F}_{\text{global}}(z').
\end{equation}

\textbf{Design rationale:} Based on the observation in Section~\ref{sec:motivation}, we use \textbf{convolution} for local processing (bounded receptive field captures fine-scale vortex structures) and \textbf{MLP-Mixer} for global processing (all-to-all spatial communication models domain-wide pressure coupling). The sequential order—local before global—ensures that fine-scale features are first extracted, then aggregated into coherent global patterns.

\subsubsection{Local Pathway: \textbf{Convolution}}

We employ depthwise separable convolutions:
\begin{equation}
\mathcal{F}_{\text{local}}(z) = z + \gamma \cdot \text{Conv}_{\text{point}}(\sigma(\text{Conv}_{\text{depth}}(\text{Norm}(z)))),
\end{equation}
where $\text{Conv}_{\text{depth}}$ is a depthwise convolution, $\text{Conv}_{\text{point}}$ is a pointwise convolution for channel mixing, and $\gamma$ is a learnable scaling parameter.

\subsubsection{Global Pathway: \textbf{MLP-Mixer}}

We employ spatial and channel mixing:
\begin{equation}
\mathcal{F}_{\text{global}}(z) = z + \text{MLP}_{\text{channel}}(\text{Norm}(\text{MLP}_{\text{spatial}}(\text{Norm}(z)))),
\end{equation}
where $\text{MLP}_{\text{spatial}}$ operates across all spatial locations for each channel, and $\text{MLP}_{\text{channel}}$ mixes feature representations across channels.

\section{Experiment}
\label{sec:Experiment}

\begin{table*}[htbp]
\centering
\caption{MSE comparison across methods for NS-Forced and NS-Decaying turbulence. ``All-step'' denotes the average error over all prediction steps, while ``$i$-step'' indicates the error at the $i$-th specific prediction step. Lower MSE values indicate better performance, while "nan" indicates excessive model error or data overflow, suggesting model collapse.}
\label{tab:main experience}
\setlength{\tabcolsep}{12pt}
\renewcommand{\arraystretch}{1.05}
\begin{tabular}{l|cccc||cccc}
\toprule
\multirow{2}{*}{Method} & 
\multicolumn{4}{c}{NS-Forced} & 
\multicolumn{4}{c}{NS-Decaying} \\
\cmidrule(r){2-5} \cmidrule(l){6-9}
& All-step & 1-step & 10-step & 19-step & 
All-step & 1-step & 50-step & 99-step \\
\midrule
\textbf{DSO}      & \textbf{0.0153} & \textbf{0.0001} & \textbf{0.0015} & \textbf{0.1086} & \textbf{0.1714} & \textbf{0.0003} & \textbf{0.1271} & \textbf{0.4986} \\
UNO         & 0.2752 & 0.0037 & 0.0974 & 1.1894 & nan    & 0.1771 & nan & nan    \\
CNO         & 0.3790 & 0.0008 & 0.1116 & 1.6521 & 4.5553 & 0.0390 & 5.2277 & 6.2792 \\
FNO         & 0.0263 & 0.0001 & 0.0024 & 0.1852 & 1.6333 & 0.0242 & 1.7299 & 2.9465 \\
U\_Net      & 0.1278 & 0.0007 & 0.0308 & 0.6620 & 2.9288 & 0.0177 & 3.3029 & 4.6783 \\
LSM         & 0.0779 & 0.0002 & 0.0086 & 0.4988 & 2.8647 & 0.0033 & 3.1483 & 5.2504 \\
Swin        & 0.1268 & 0.0002 & 0.0271 & 0.6362 & 3.2328 & 0.0070 & 3.7928 & 5.0429 \\
ConvLSTM    & 1.3521 & 0.0101 & 1.2379 & 3.1692 & 3.3228 & 0.1078 & 3.7786 & 3.1913 \\
SimVP       & 0.0718 & 0.0003 & 0.0115 & 0.4290 & 1.5510 & 0.0016 & 1.5575 & 3.2480 \\
ResNet      & 0.5906 & 0.0023 & 0.2180 & 2.1835 & 4.7140 & 0.0674 & 5.3391 & 6.6097 \\
PastNet     & 0.1349 & 0.0012 & 0.0326 & 0.6009 & 3.2089 & 0.1879 & 3.6807 & 5.1670 \\
\bottomrule
\end{tabular}
\end{table*}


\subsection{Experiment Setting}
\paragraph{Implementation Details}
All models in this paper are trained on a server equipped with eight NVIDIA A100 GPUs (40GB memory per card). We employ the DistributedDataParallel mode from the PyTorch framework~\cite{paszke2019pytorch} for distributed training to accelerate convergence. Model inference is conducted on a single NVIDIA A100 GPU. The experimental software environment is built on Python 3.8, with PyTorch 1.8.1 (CUDA 11.1) and TorchVision 0.9.1. Throughout training, we maintain a consistent hyperparameter configuration: batch size of 20, total epochs of 500, and an initial learning rate of 0.001. To ensure experimental reproducibility, all random seeds are fixed to 42. For training, the model adopts one-step prediction, i.e., using one time step to predict the next. During testing, a rollout approach is used: the first time step predicted the next, and the result was fed as input for subsequent predictions, iterating this cycle to forecast multiple time steps. Further experimental details are provided in the Appendix \ref{app: Implementation Details}.
\paragraph{Datasets}
We evaluate various models on two representative fluid dynamics datasets, aiming to assess their ability to capture long-term fluid motion under different physical conditions.

\begin{itemize}
    \item \textbf{NS-Forced (Navier-Stokes Forced Turbulence) Benchmark}: 
This dataset is based on the Navier-Stokes benchmark proposed by~\cite{li2021fourier}. The kinematic viscosity is set to \(\nu = 10^{-5}\). The initial vorticity field \(\omega_0\) is sampled from a zero-mean Gaussian random field with a covariance operator \((-\Delta + \tau^2 I)^{-\alpha/2}\), where \(\Delta\) is the Laplacian operator, \(I\) is the identity operator, and \(\tau\) and \(\alpha\) are parameters controlling the shape of the energy spectrum. The energy density \(E(k)\) follows the decay law \(E(k) \sim (k^2 + \tau^2)^{-\alpha}\). The flow is continuously driven by a fixed external force at low wavenumbers, with no additional drag effects introduced.
    \item \textbf{NS-Decaying (Navier-Stokes Decaying Turbulence)}:
This dataset simulates the natural decay of turbulence~\cite{mcwilliams1984emergence}. The power spectrum of the initial stream function is defined by \(\hat{\psi}(k)^2 \sim k^{-1} (\tau_0^2 + (k/k_0)^4)^{-1}\), where \(k_0\) is the characteristic wavenumber and \(\tau_0\) is a spectral shape parameter; this expression determines the structure of the initial flow field. This initial condition is specifically designed to achieve slow energy dissipation, and the evolution of vorticity density over time exhibits characteristics resembling the Kolmogorov energy cascade. This turbulence is force-free and evolves entirely from the initial conditions.
\end{itemize}
More details can be found in the Appendix \ref{app: Datasets Details}.

\paragraph{Baseline}
We evaluate our method against leading models from neural operators, computer vision, and time series prediction. For neural operators, we select FNO \cite{li2021fourier}, UNO \cite{rahman2022u}, CNO \cite{raonic2023convolutional}, and LSM \cite{pmlr-v202-wu23f} as representative architectures for learning PDE solution mappings. In computer vision, we adopt ResNet \cite{he2016deep}, U-net \cite{ronneberger2015u}, and Swin Transformer \cite{liu2021swin} due to their strong spatial feature extraction capabilities. For time series prediction, we employ ConvLSTM \cite{lin2020self}, PastNet \cite{wu2024pastnet}, and SimVP \cite{gao2022simvp}, which excel in capturing temporal dynamics. The model sizes and configurations are detailed in Appendix \ref{app: Model Parameters}.

\subsection{Main Results}
\label{sec: main results}
\begin{figure*}[htbp]
\centering
\includegraphics[width=0.95\textwidth, height=0.58\textheight]{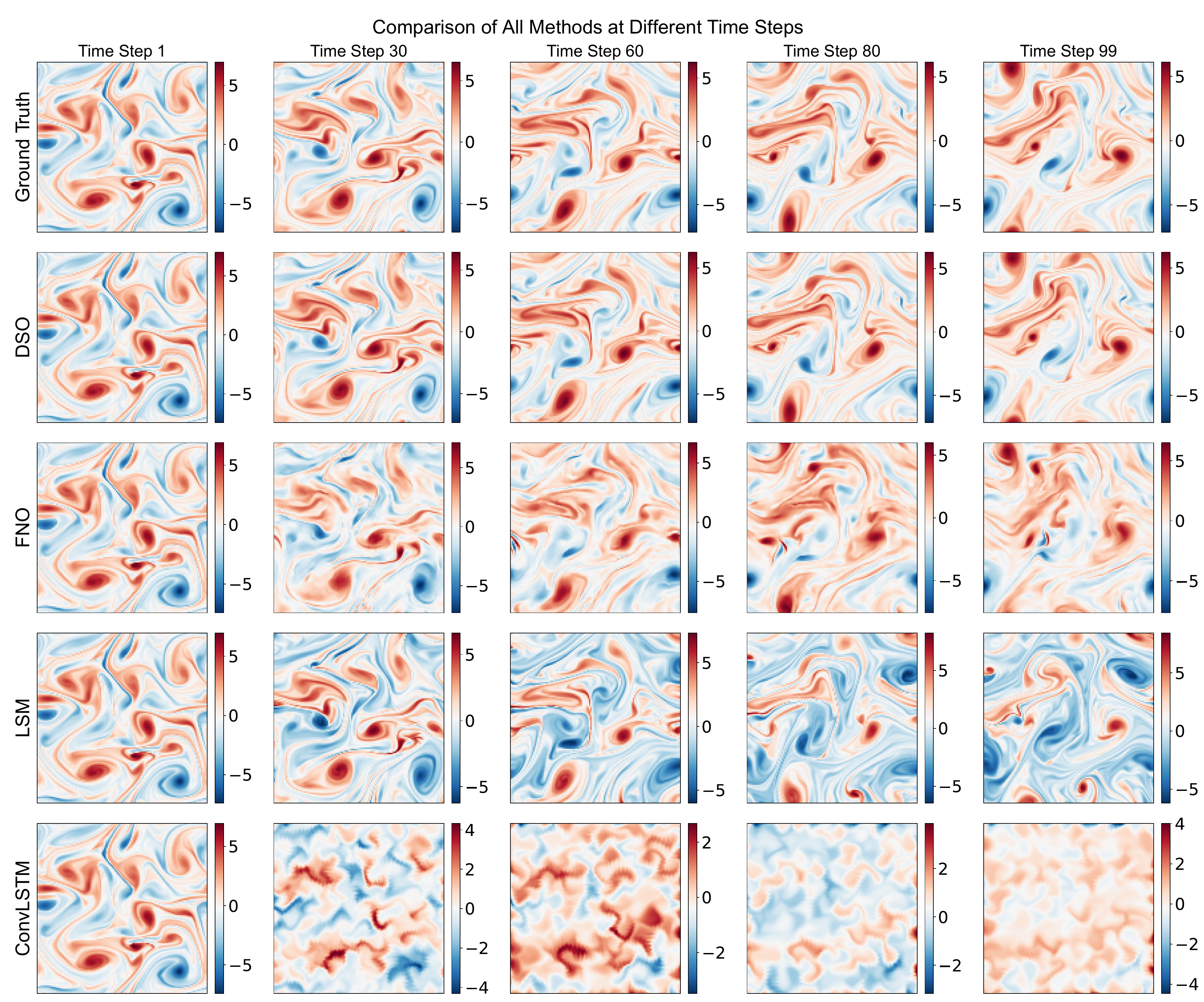}
\caption{Prediction visualization of selected models across various time steps for the NS-Decaying case, where the top row displays the ground truth values.}
\label{fig:main experience}
\end{figure*}

Table \ref{tab:main experience} provides a comprehensive comparison of the Mean Squared Error (MSE) performance for various methods across two scenarios: NS-Forced and NS-Decaying turbulence. Meanwhile, Figure \ref{fig:main experience} showcases the prediction results for a subset of these methods, with the visualizations for the remaining methods detailed in the Appendix \ref{app: Visualization Results}. Our DSO approach demonstrates superior performance and marked long-term stability, which is substantiated by the following key observations:

\textbf{Consistent State-of-the-Art Performance}: DSO achieves the lowest error across nearly all evaluation metrics in both datasets. For NS-Forced, DSO reduces the all-step error to just 58.2\% of the second-best method's performance (0.0153 vs. FNO's 0.0263). In the more challenging NS-Decaying scenario, DSO achieves an even more dramatic improvement, reducing the all-step error to merely 11.0\% of the next best performer's result (0.1714 vs. SimVP's 1.5510), which means our method reduces the prediction error by over 88\%.

\textbf{Exceptional Long-Term Stability}: DSO demonstrates remarkable robustness in extended temporal predictions, maintaining physical consistency where competing methods catastrophically fail. In NS-Forced at the critical 19-step horizon (near simulation limits), DSO sustains a remarkably low error of 0.1086—reducing the error to merely 58.6\% of FNO's performance (0.1852), 9.1\% of UNO's (1.1894), 6.6\% of CNO's (1.6521), and 5.0\% of ResNet's (2.1835). 
The stability advantage becomes even more pronounced in NS-Decaying at the extreme 99-step horizon. While most methods suffer numerical collapse (UNO produces NaN values across all long-horizon metrics), DSO achieves an error of 0.4986—reducing the error to just 16.9\% of FNO's result (2.9465), 15.4\% of SimVP's (3.2480).  

\textbf{Visualization Analysis}: Figure \ref{fig:main experience} presents the experimental results of selected methods on the NS-Decaying dataset. The visualization reveals several limitations in baseline approaches: the best-performing baseline method FNO exhibits blurred turbulent details; ConvLSTM suffers from numerical collapse during long-term predictions; While LSM maintains detailed flow structures, it demonstrates significant motion drift over extended horizons. For instance, LSM's prediction incorrectly splits the vortex in the central-lower region into two structures. In contrast, our method successfully avoids all these issues. Even at the 99th time step, our predictions remain visually nearly identical to the ground truth.

\begin{figure}[htbp]
\centering
\includegraphics[width=0.5\textwidth]{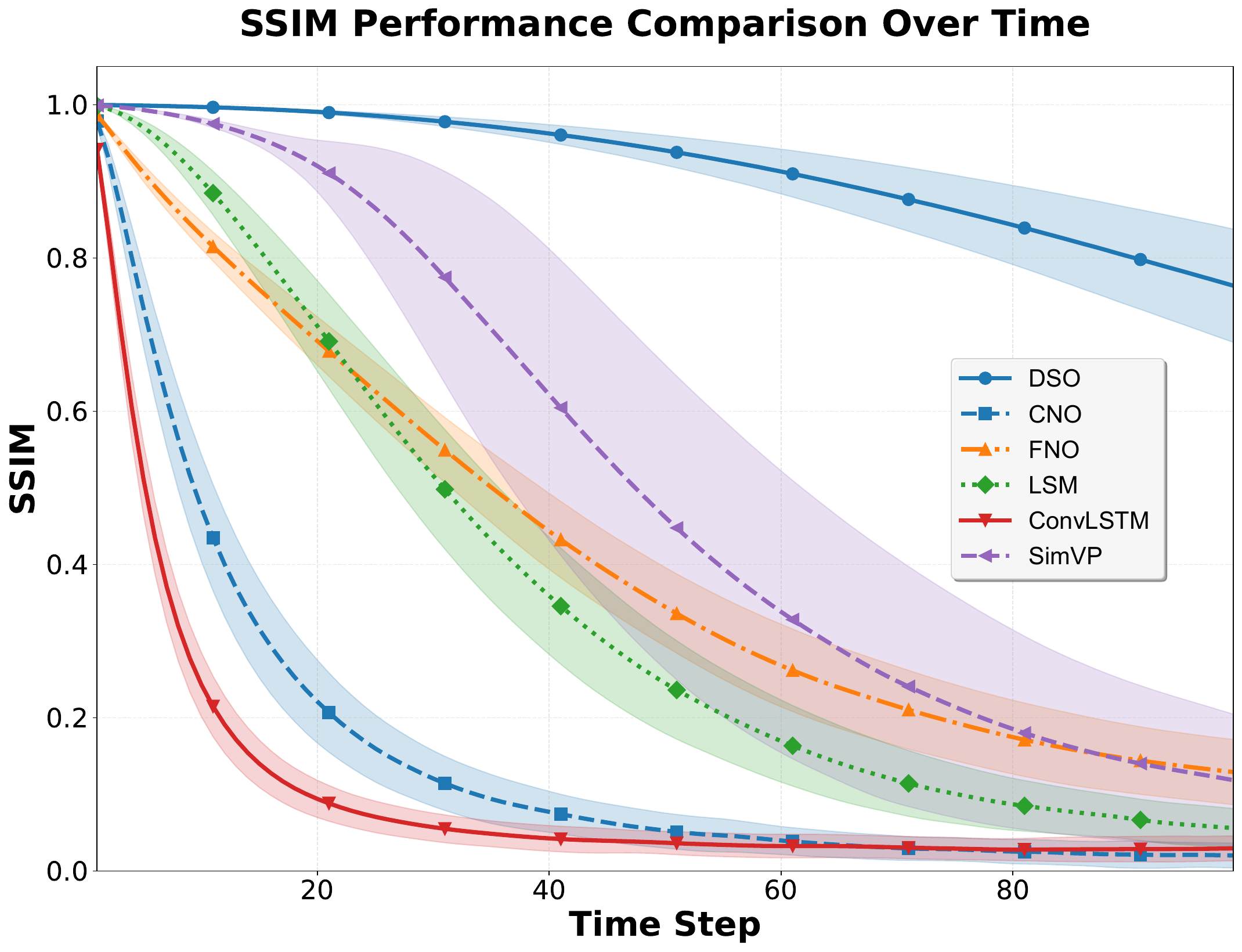}
\caption{We selected predictions from selected models on the NS-Decaying dataset, computed and analyzed the mean and variance of their SSIM values.}
\label{fig:ssim}
\vspace{-13pt}
\end{figure}

This stable error propagation—devoid of numerical instabilities even at the maximum prediction horizon—validates the core innovation of DSO: its dual-scale architecture explicitly separates local and global processing to match the fundamental dichotomy in fluid dynamics. As demonstrated in our motivation experiments (Section~\ref{sec:motivation}), local information primarily governs fine-scale vortex structure and gradient dynamics, while global information controls motion trajectories through domain-wide pressure coupling. By employing depthwise separable convolutions for local feature extraction (capturing gradient intensification and vortex deformation) and MLP-Mixer for global information aggregation (modeling long-range spatial dependencies), DSO simultaneously preserves fine-grained turbulent details in regions with steep gradients while maintaining coherent large-scale motion predictions. This physics-informed dual-scale design meets the critical requirements for reliable fluid prediction in scientific and engineering applications—especially in scenarios requiring accurate long-horizon forecasting where single-scale approaches fail to capture the multi-scale nature of turbulent flows.


\subsection{Analytical Experiment}
\subsubsection{SSIM Evaluation Metrics}
In fluid dynamics forecasting, relying solely on Mean Squared Error (MSE) is limited. MSE measures only pixel-level numerical discrepancies and fails to distinguish between critical physical structure errors (like vortex displacement) and minor numerical fluctuations. Given that turbulence prediction prioritizes the preservation of critical physical features such as vortex morphology, we complement our analysis with the Structural Similarity Index (SSIM) (details provided in the Appendix \ref{app: Evaluation Indicators}).

Our experimental results show that DSO maintains significant superiority in SSIM metrics, particularly ensuring stable forecasting reliability during long-term predictions. We also note that while certain baselines (SimVP, LSM, FNO) show acceptable MSE performance mid-term ($\sim 50$ time steps), their structural preservation varies. For instance, LSM and ConvLSTM may exhibit similar MSE at 50 steps, but Figure \ref{fig:ssim} clearly shows LSM's superior ability to preserve discernible vortex features. This comparison underscores the critical importance of evaluating both quantitative metrics (MSE) and structural fidelity (SSIM) in assessing turbulence forecasting performance.

\subsubsection{Gradient and Divergence Evaluation Metrics}
Gradient $\nabla \mathbf{u} = \left(\frac{\partial u}{\partial x}, \frac{\partial u}{\partial y}\right)$ and divergence $\nabla \cdot \mathbf{u} = \frac{\partial u}{\partial x} + \frac{\partial v}{\partial y}$ analysis validates physical fidelity of predictions. Gradient fields capture critical deformation features—strain rates, vortex stretching, and coherent structures—that govern turbulent energy transfer. Divergence fields verify mass conservation, where non-zero values reveal unphysical instabilities. Together, they ensure predictions maintain both structural accuracy and fundamental physical constraints, essential for trustworthy scientific applications.

Experimental results demonstrate that DSO achieves significantly lower prediction errors than competing methods, demonstrating exceptional capability in capturing flow dynamics and maintaining dynamic consistency across all prediction horizons.

\subsection{Ablations}
To analyze the contribution of individual modules in our dual-scale architecture, we conduct ablation studies by separately removing the convolution module (local processing) and the MLP-Mixer module (global processing) from the MixBlock, while keeping all other hyperparameters and settings identical to the main experiments. Specifically, ``w/o Conv'' removes the depthwise separable convolution branch, and ``w/o MLP-Mixer'' removes the MLP-Mixer branch.

\begin{table}[htbp]
\centering
\caption{Ablation study of DSO on the NS-Decaying dataset}
\label{tab:ablation}
\renewcommand{\arraystretch}{1}
\begin{tabular}{l|cccc}
\toprule
Method & All-step & 1-step & 50-step & 99-step \\
\midrule
DSO & 0.1714 & 0.0003 & 0.1271 & 0.4986 \\
w/o Conv & 0.2059 & 0.0004 & 0.1565 & 0.5871 \\
w/o MLP-Mixer & 1.1335 & 0.0010 & 1.1330 & 2.4606 \\
\bottomrule
\end{tabular}
\end{table}

The results reveal distinct roles for each module in our dual-scale design. Removing the MLP-Mixer module causes the most significant performance degradation (All-step error increases from 0.1714 to 1.1335, a 6.6$\times$ increase), confirming that global processing is essential for capturing long-range spatial dependencies and maintaining prediction stability. At the 99-step horizon, the error increases nearly 5$\times$ (from 0.4986 to 2.4606), demonstrating that without global information aggregation, the model fails to accurately predict large-scale motion trajectories. Removing the convolution module also degrades performance (All-step error increases to 0.2059), indicating that local processing contributes to capturing fine-scale turbulent structures. These results validate our dual-scale design: both local and global processing are indispensable, with global processing playing a particularly critical role in long-term prediction stability.


\begin{figure*}[htbp]
\centering
\begin{subfigure}[b]{0.45\textwidth}
    \centering
    \includegraphics[width=\linewidth, height=5.3cm]{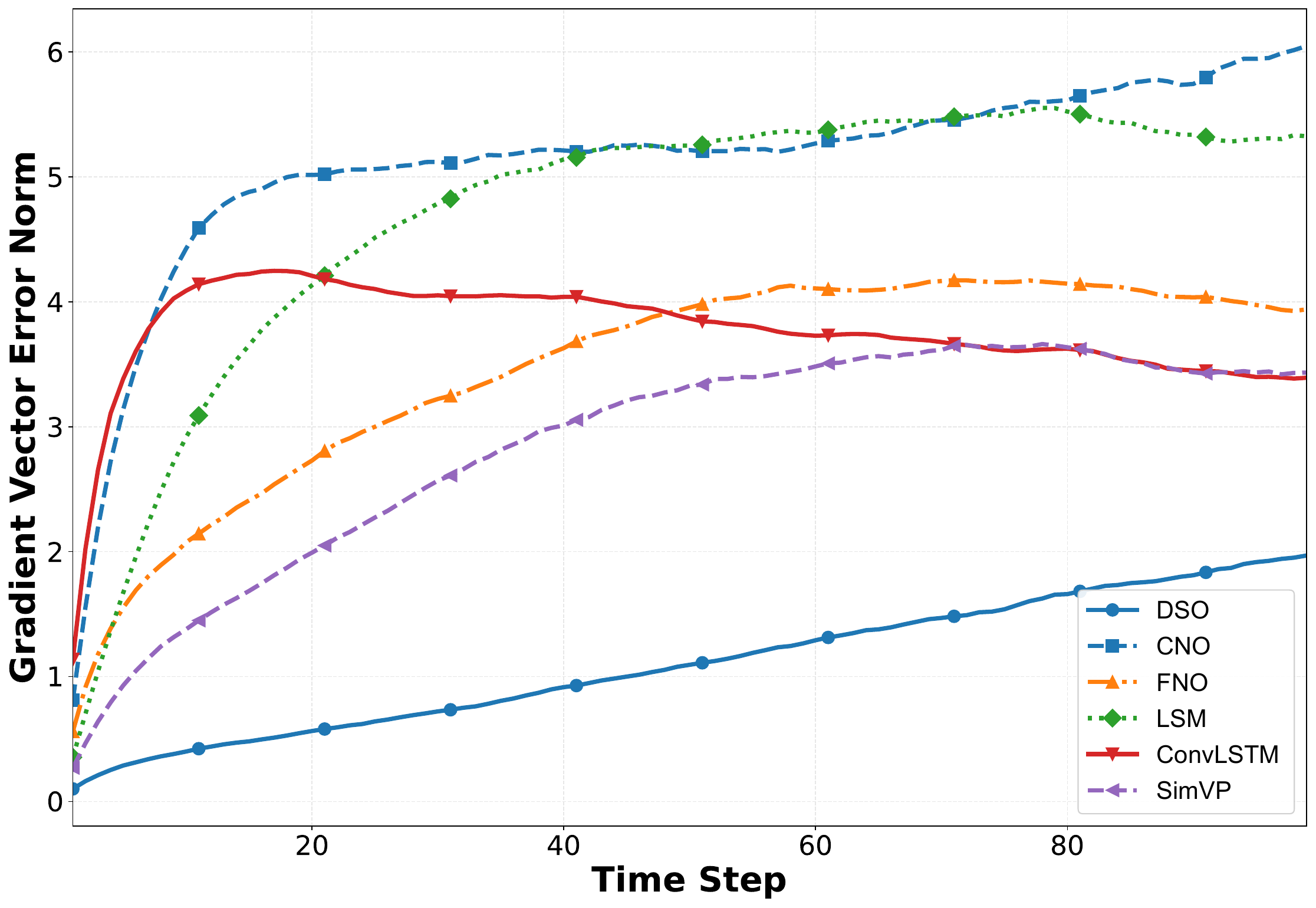}
    \caption{Gradient Vector error comparison over time}
    \label{fig:gradient}
\end{subfigure}
\hfill
\begin{subfigure}[b]{0.45\textwidth}
    \centering
    \includegraphics[width=\linewidth, height=5.3cm]{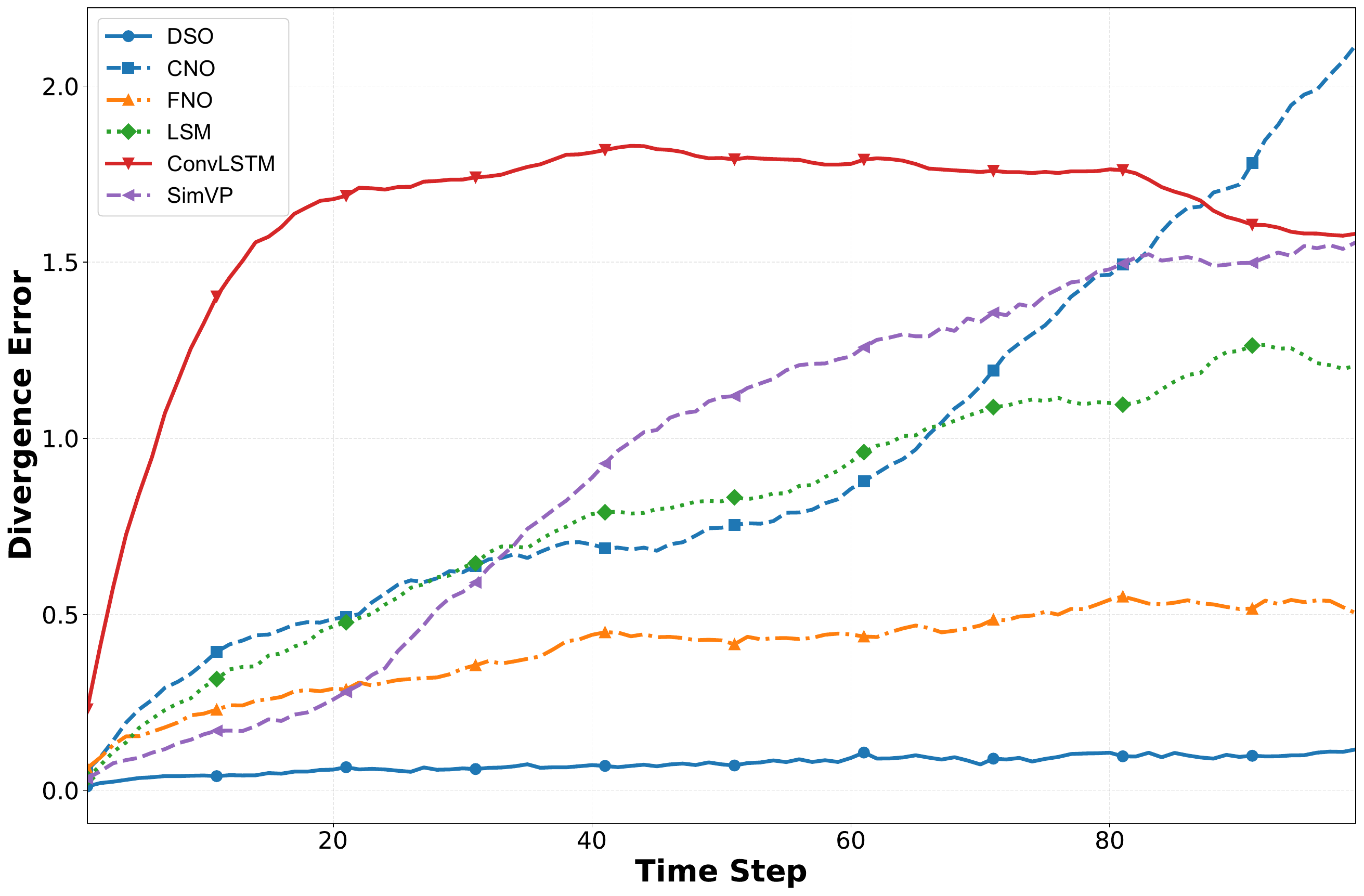}
    \caption{Divergence error comparison over time}
    \label{fig:divergence}
\end{subfigure}
\caption{Comparison of gradient and divergence errors of different methods on the NS-Decaying dataset. (a) Gradient magnitude error quantifies the accuracy of capturing flow deformation features. (b) Divergence error measures physical consistency and mass conservation properties.}
\label{fig:grad_div_comparison}
\end{figure*}

\section{Conclusion}

We propose DSO (Dual-scale Spatiotemporal Operator), a neural operator that combines local convolutions and global MLP-Mixer to capture multi-scale flow dynamics. Experiments on forced and decaying turbulence benchmarks demonstrate that DSO addresses key limitations of existing methods—blurred details, trend bias, and numerical collapse—achieving up to 89\% error reduction while maintaining stability at extreme time horizons. These results establish DSO as a robust solution for long-horizon fluid dynamics prediction.

\section{Limitations and Ethical Considerations}

While our method shows promising results, future work may explore extensions to broader physical scenarios and boundary conditions. This work focuses on scientific computing for fluid dynamics, posing minimal ethical risks. We encourage responsible use and proper validation in downstream applications.

\clearpage
\normalem
\bibliographystyle{ACM-Reference-Format}
\bibliography{sample-base}

\appendix
\section{Appendix}

\subsection{Motivation Experiment Details}
\label{app:motivation_details}

\subsubsection{Governing Equations and Numerical Method}

We simulate 2D incompressible flow governed by the vorticity-streamfunction formulation of the Navier-Stokes equations:
\begin{equation}
    \frac{\partial \omega}{\partial t} + (\mathbf{u} \cdot \nabla) \omega = \nu \nabla^2 \omega, \quad \nabla^2 \psi = -\omega
\end{equation}
where $\omega$ is the vorticity, $\psi$ is the streamfunction, $\mathbf{u} = (\partial_y \psi, -\partial_x \psi)$ is the velocity, and $\nu$ is the viscosity. We use a pseudo-spectral method with 4th-order Runge-Kutta time integration on a $128 \times 128$ periodic domain.

\subsubsection{Experimental Configuration}

We construct two scenarios to isolate local and global effects:
\begin{itemize}[leftmargin=*,nosep]
    \item \textbf{Scenario A (Close Perturbation)}: A vortex dipole (counter-rotating pair) with a single perturbing vortex placed at distance $d_{close} = 0.6$ from the dipole center.
    \item \textbf{Scenario B (Far Perturbation)}: The same dipole configuration with the perturbing vortex placed at distance $d_{far} = 2.5$.
\end{itemize}
The dipole is identical in both cases, allowing direct comparison of how perturbation distance affects the dynamics.

\subsubsection{Quantitative Metrics}

To quantify the observations, we track two metrics:
\begin{itemize}[leftmargin=*,nosep]
    \item \textbf{Local Deformation}: Maximum vorticity gradient $\|\nabla \omega\|_{\max}$ within a local region centered at the vortex center, measuring fine-scale structure intensity.
    \item \textbf{Global Displacement}: Center-of-vorticity position 
    $$\mathbf{x}_c = \int \mathbf{x} |\omega| d\mathbf{x} / \int |\omega| d\mathbf{x}$$.
\end{itemize}

Table~\ref{tab:motivation_appendix} summarizes the results:

\begin{table}[h]
\centering
\caption{Comparison of local and global effects for close vs. far perturbations.}
\label{tab:motivation_appendix}
\begin{tabular}{lcc}
\toprule
Metric & Close ($d=0.6$) & Far ($d=2.5$) \\
\midrule
Local gradient change $\Delta\|\nabla\omega\|_{\max}$ & \textbf{+45\%} & $-$29\% \\
Global position shift $\|\Delta \mathbf{x}_c\|$ & 0.8 & 0.6 \\
\bottomrule
\end{tabular}
\end{table}

The key distinction lies in the \textit{nature} of the effect: close perturbation causes gradient \textit{increase} due to vortex stretching and fine-scale interactions, while far perturbation leads to gradient \textit{decay} as the dipole evolves undisturbed. Both scenarios show similar global displacement magnitudes, but the underlying mechanisms differ fundamentally.

\subsection{Experiment Details}
\subsubsection{Implementation Details}
\label{app: Implementation Details}

Specifically, the dataset is partitioned into training, validation, and testing sets following an $80\%:10\%:10\%$ ratio. During the training phase, we adopt a one-step-ahead prediction strategy, utilizing the state at the current time step to predict the state at the next. Taking the NS-Decaying dataset as an example, a single trajectory comprises 100 continuous time steps, which allows for the construction of 99 one-step prediction tasks (where each pair of adjacent time steps constitutes a training sample). For the testing phase, we employ a rolling prediction scheme: based on the initial time step, the model predicts the second step, and this predicted output is then recursively used as the input for subsequent steps, ultimately achieving multi-step prediction across the entire horizon.

\subsubsection{Datasets Details}
\label{app: Datasets Details}

We test our method, on two standard datasets for 2D isotropic turbulence. We chose these specific cases to see if the model can handle chaotic fluid dynamics in two different situations: one with external forcing and one with natural decay.
\paragraph{NS-Forced (Navier-Stokes Forced Turbulence).}
\begin{itemize}
    \item \textbf{NS-Forced}. This dataset is based on the Navier-Stokes equation benchmark introduced by Li et al. \cite{li2021fourier} in their work on Fourier Neural Operators (FNO). It simulates the evolution of vorticity in a 2D viscous, incompressible fluid on the unit torus $(0, 1)^2$. The governing equations in vorticity form are:

\begin{align}
    &\begin{aligned}
    \frac{\partial w(\mathbf{x}, t)}{\partial t} + \mathbf{u}(\mathbf{x}, t) \cdot \nabla w(\mathbf{x}, t) &= \nu \Delta w(\mathbf{x}, t) + f(\mathbf{x}), \\
    &\quad \forall \mathbf{x} \in (0, 1)^2, t \in (0, T]
    \end{aligned} \\[0.5ex]
    &\nabla \cdot \mathbf{u}(\mathbf{x}, t) = 0 \quad \forall \mathbf{x} \in (0, 1)^2, t \in [0, T] \\[0.5ex]
    &w(\mathbf{x}, 0) = w_0(\mathbf{x}) \quad \forall \mathbf{x} \in (0, 1)^2
\end{align}

\noindent Here, $w$ represents the vorticity field (defined as $w = \nabla \times \mathbf{u}$), while $\mathbf{u}$ denotes the velocity field. The term $w_0$ refers to the initial vorticity, $\nu$ is the kinematic viscosity, and $f(\mathbf{x})$ is a constant external force.

Following the setup by Li et al. [19], we set the kinematic viscosity to $\nu = 1 \times 10^{-5}$. To create the initial vorticity $w_0$, we sample from a zero-mean Gaussian random field. This field uses a covariance operator defined as $(-\Delta + \tau^2 I)^{-\alpha/2}$, where $\Delta$ is the Laplacian and $I$ is the identity operator. The parameters $\tau$ and $\alpha$ shape the energy spectrum density $E(k)$ so that $E(k) \sim (k^2 + \tau^2)^{-\alpha}$. The external force $f(\mathbf{x})$ focuses mainly on low wavenumbers to keep adding energy to the system, and we do not include any drag effects.

This dataset is used to evaluate the model's long-term prediction capability and its ability to capture chaotic dynamic behaviors under continuous external energy input. In our experiments, the data shape is $(1200, 20, 1, 64, 64)$, indicating that the dataset contains 1200 samples, each with 20 time steps, a feature dimension of 1, and a spatial resolution of $64\times64$. For each sample, we extend the prediction horizon to 19 time steps.
\item \textbf{NS-Decaying (Navier-Stokes Decaying Turbulence)} \cite{mcwilliams1984emergence}.

For the second benchmark, we simulate the classic case of 2D decaying turbulence, adhering to the seminal initialization protocols established by McWilliams \cite{mcwilliams1984emergence} in ``The emergence of isolated coherent vortices in turbulent flow.'' A key distinction from NS-Forced is the absence of external forcing after the initial perturbation; consequently, the system's energy dissipates naturally over time.

To characterize the initial flow field, we utilize the power spectrum of the streamfunction, $\psi(\mathbf{x})$. The distribution of the squared magnitude of the Fourier coefficients, denoted as $|\hat{\psi}(\mathbf{k})|^2$ with wavevector $\mathbf{k}$, is given by:

\begin{equation}
    |\hat{\psi}(\mathbf{k})|^2 \sim k^{-1}(\tau_0^2 + (k/k_0)^4)^{-1} \tag{A.35}
\end{equation}

\noindent where $k = |\mathbf{k}|$ is the wavenumber magnitude, and the parameters $\tau_0$ and $k_0$ dictate the shape of the spectrum. This specific configuration is designed to drive slow energy decay while enabling the flow to self-organize into coherent vortex structures. During this evolution, the enstrophy density spectrum develops features comparable to a Kolmogorov energy cascade.

We employ this dataset to rigorously assess the model's fidelity and stability within a self-evolving system, specifically focusing on its capacity to capture long-term dynamics, energy dissipation, and complex structural evolution. In our experiments, the data shape is $(1200, 100, 1, 128, 128)$, indicating that the dataset contains 1200 samples, each with 100 time steps, a feature dimension of 1, and a spatial resolution of $128\times128$. For each sample, we extend the prediction horizon to 99 time steps.
\end{itemize}

\subsubsection{Model Parameters}
\label{app: Model Parameters}
Below, we introduce the core parameters our method uses for different datasets. The model hyperparameters for the NS-Forced dataset are listed in Table~\ref{tab:model_params_NS-Forced}, and those for the NS-Decaying dataset are listed in Table~\ref{tab:model_params_NS-Decaying}.

\subsubsection{Visualization Results}
\label{app: Visualization Results}
\begin{figure*}[htbp]
\centering
\includegraphics[width=0.8\textwidth, height=1\textheight]{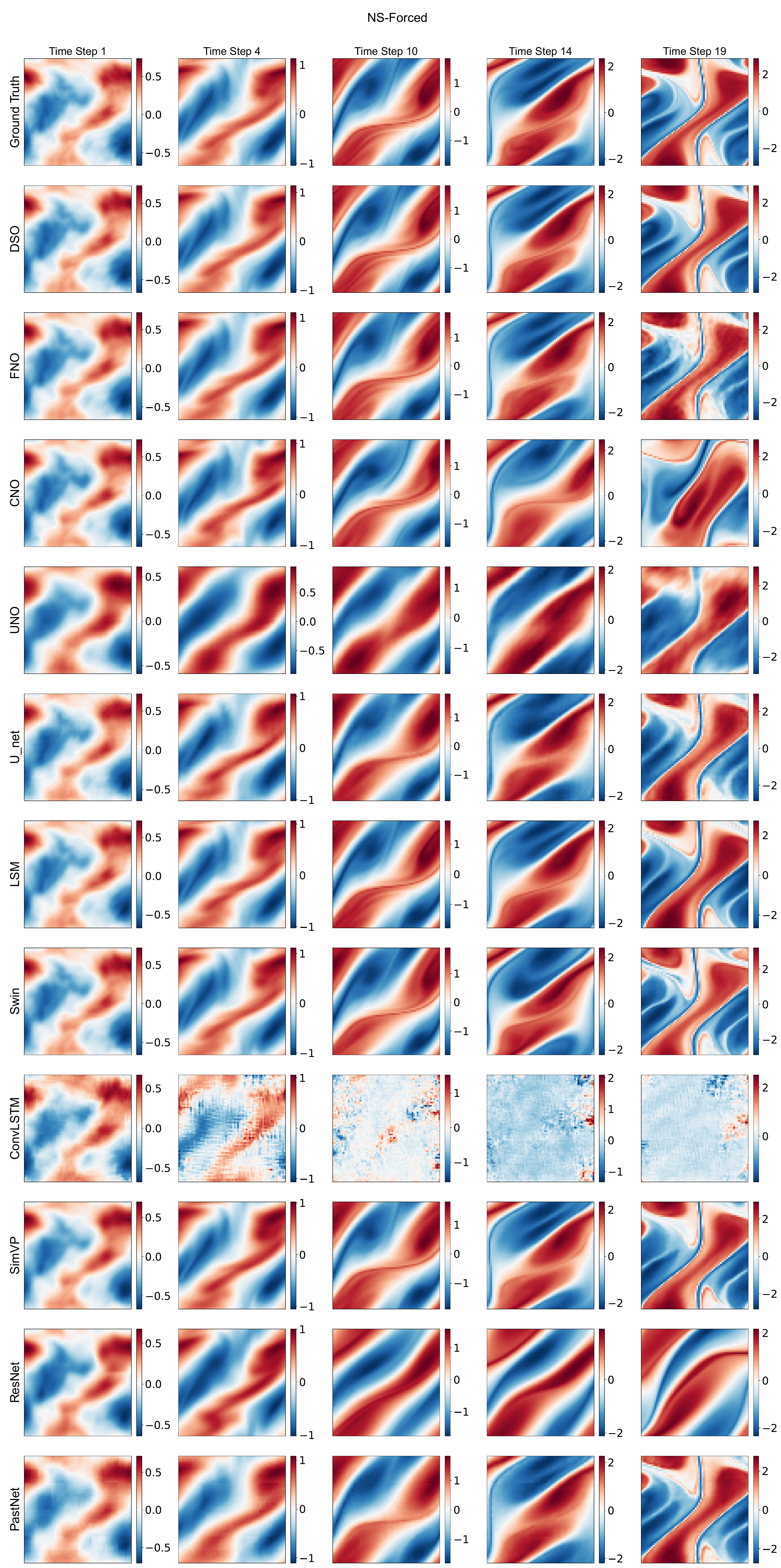}
\caption{Prediction visualization of selected models across various time steps for the NS-Forced case, where the top row displays the ground truth values.}
\label{fig:all_Forced}
\end{figure*}

\begin{figure*}[htbp]
\centering
\includegraphics[width=0.8\textwidth, height=1\textheight]{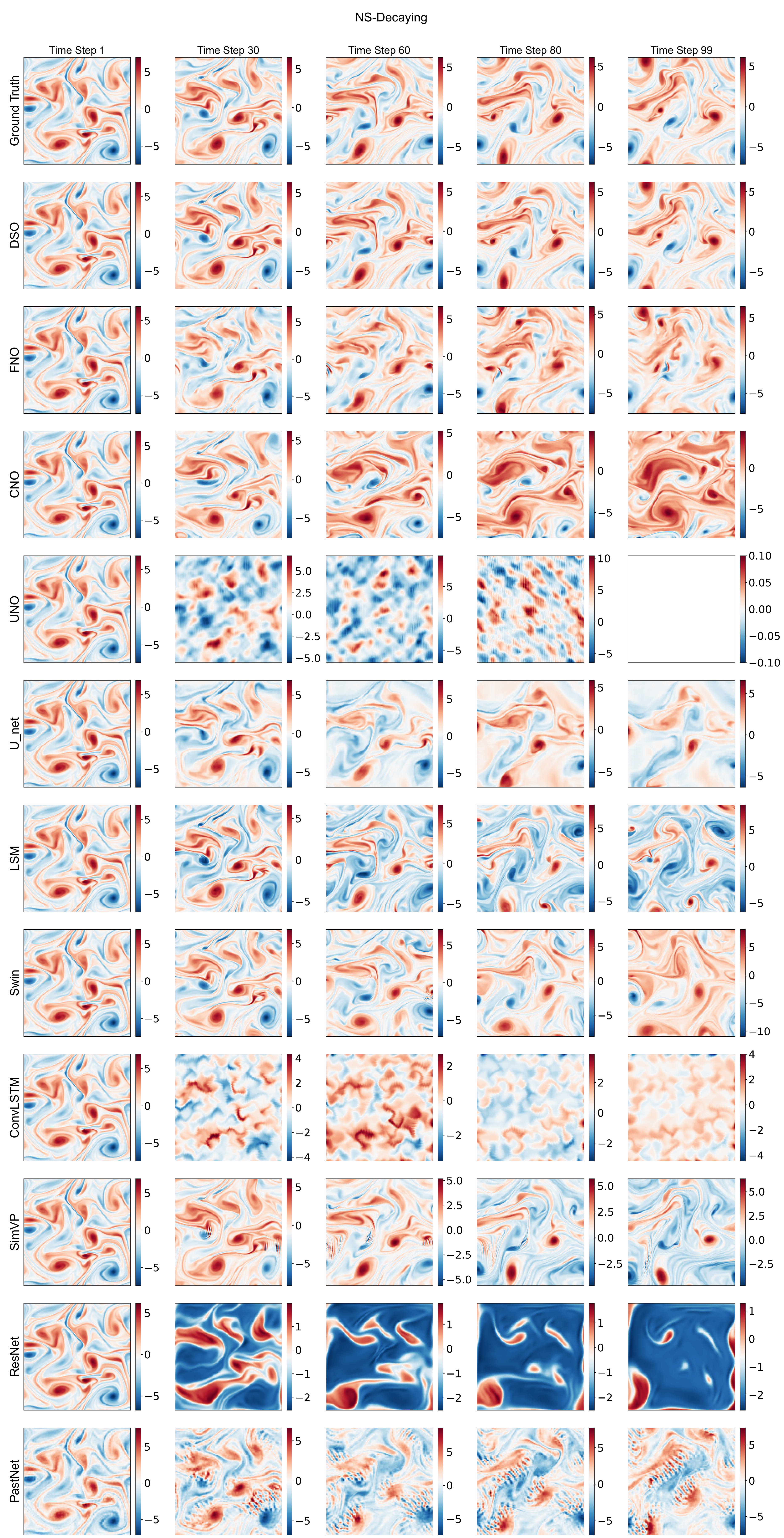}
\caption{Prediction visualization of selected models across various time steps for the NS-Decaying case, where the top row displays the ground truth values.}
\label{fig:all_Decaying}
\end{figure*}

We present the visualization results of all models from the main experiments in Section~\ref{sec: main results}. Figure~\ref{fig:all_Forced} shows the prediction visualization on the NS-Forced dataset, and Figure~\ref{fig:all_Decaying} shows the visualization on the NS-Decaying dataset.

\subsubsection{Evaluation Indicators}
\label{app: Evaluation Indicators}

\paragraph{MSE}
Mean Squared Error (MSE) is a fundamental metric used to quantify the average squared difference between predicted values and actual observed values. The specific form is as follows:
\begin{align}
    \text{MSE} = \frac{1}{n} \sum_{i=1}^{n} (y_i - \hat{y}_i)^2,
\end{align}

\paragraph{SSIM}
Structural Similarity Index Measure (SSIM) is a perceptual metric used to quantify the image quality degradation between a reference image and a processed image. Unlike traditional error metrics, SSIM considers human visual perception by evaluating luminance, contrast, and structure comparisons. The specific form is as follows:
\begin{align}
    \text{SSIM}(x,y) = \frac{(2\mu_x\mu_y + C_1)(2\sigma_{xy} + C_2)}{(\mu_x^2 + \mu_y^2 + C_1)(\sigma_x^2 + \sigma_y^2 + C_2)},
\end{align}
where $\mu_x$, $\mu_y$ are the local means, $\sigma_x$, $\sigma_y$ are the standard deviations, and $\sigma_{xy}$ is the cross-covariance between images $x$ and $y$. Constants $C_1$, $C_2$ stabilize the division.

\clearpage

\begin{table}[htbp]
\centering
\caption{Model Parameters Configuration of NS-Forced}
\label{tab:model_params_NS-Forced}
\begin{tabular}{lll}
\toprule
\textbf{Model Name} & \textbf{Parameter} & \textbf{Value(s)} \\
\midrule
DSO & Input Shape & [1,1,64,64] \\
        & Encode Hidden Dim & 128 \\
        & Translator Hidden Dim & 256 \\
        & Encode Layers & 4 \\
        & Translator Layers & 8 \\
\midrule
UNO & in\_width & 5 \\
    & width & 32 \\
\midrule
CNO & Input Channels (in\_dim) & 1 \\
    & Input Size (H,W) & 64,64 \\
    & Num Layers (N\_layers) & 4 \\
    & Channel Multiplier & 32 \\
    & Latent Lift Proj Dim & 64 \\
    & Activation & 'cno\_lrelu' \\
\midrule    
FNO & Modes (modes1, modes2) & 16, 16 \\
    & Width & 64 \\
    & Input Channels & 1 \\
    & Output Channels & 1 \\
\midrule
U\_Net & Input Channels & 1 \\
       & Output Channels & 1 \\
       & Kernel Size & 3 \\
       & Dropout Rate & 0.5 \\
\midrule
LSM & Model Dim (d\_model) & 64 \\
    & Num Tokens & 4 \\
    & Num Basis & 16 \\
    & Patch Size & [4,4] \\
\midrule
Swin & Hidden Dim & 128\\
& Input Resolution & [64,64]\\
\midrule
ConvLSTM & Input Dim & 1\\
& Hidden Dims & [64, 64]\\
& Num Layers & 2 \\
\midrule
SimVP & Input Shape & [1,1,64,64] \\
      & Hidden Dim S (hid\_S) & 128 \\
      & Hidden Dim T (hid\_T) & 256 \\
      & Output Dim & 1 \\
\midrule
ResNet  & Input & 1\\
        & Output & 1\\
\midrule
PastNet & Input Shape (T,C,H,W) & [1, 1, 64, 64]\\
& Hidden Dim T & 256 \\
&Inception Kernels &[3, 5, 7, 11]\\
\bottomrule
\end{tabular}
\end{table}

\begin{table}[htbp]
\centering
\caption{Model Parameters Configuration of NS-Decaying}
\label{tab:model_params_NS-Decaying}
\begin{tabular}{lll}
\toprule
\textbf{Model Name} & \textbf{Parameter} & \textbf{Value(s)} \\
\midrule
DSO & Input Shape & [1,1,128,128] \\
        & Encode Hidden Dim & 128 \\
        & Translator Hidden Dim & 256 \\
        & Encode Layers & 4 \\
        & Translator Layers & 8 \\
\midrule
UNO & in\_width & 5 \\
    & width & 32 \\
\midrule
CNO & Input Channels (in\_dim) & 1 \\
    & Input Size (H,W) & 128,128 \\
    & Num Layers (N\_layers) & 4 \\
    & Channel Multiplier & 32 \\
    & Latent Lift Proj Dim & 64 \\
    & Activation & 'cno\_lrelu' \\
\midrule    
FNO & Modes (modes1, modes2) & 16, 16 \\
    & Width & 64 \\
    & Input Channels & 1 \\
    & Output Channels & 1 \\
\midrule
U\_Net & Input Channels & 1 \\
       & Output Channels & 1 \\
       & Kernel Size & 3 \\
       & Dropout Rate & 0.5 \\
\midrule
LSM & Model Dim (d\_model) & 64 \\
    & Num Tokens & 4 \\
    & Num Basis & 16 \\
    & Patch Size & [4,4] \\
\midrule
Swin & Hidden Dim & 128\\
& Input Resolution & [128,128]\\
\midrule
ConvLSTM & Input Dim & 1\\
& Hidden Dims & [64, 64]\\
& Num Layers & 2 \\
\midrule
SimVP & Input Shape & [1,1,128,128] \\
      & Hidden Dim S (hid\_S) & 128 \\
      & Hidden Dim T (hid\_T) & 256 \\
      & Output Dim & 1 \\
\midrule
ResNet  & Input & 1\\
        & Output & 1\\
\midrule
PastNet & Input Shape (T,C,H,W) & [1,1,128,128]\\
& Hidden Dim T & 256 \\
&Inception Kernels &[3, 5, 7, 11]\\
\bottomrule
\end{tabular}
\end{table}

\end{document}